\documentclass[10pt,twocolumn,letterpaper]{article}

\usepackage[pagenumbers]{wacv} 

\usepackage[accsupp]{axessibility} 

\usepackage{graphicx}
\usepackage{amsmath}
\usepackage{amssymb}
\usepackage{booktabs}
\usepackage{pifont}
\usepackage{multirow}
\usepackage{comment}
\usepackage{xcolor}
\usepackage{placeins}
\usepackage{appendix}

\usepackage[pagebackref,breaklinks,colorlinks]{hyperref}
\usepackage[capitalize]{cleveref}
\crefname{section}{Sec.}{Secs.}
\Crefname{section}{Section}{Sections}
\Crefname{table}{Table}{Tables}
\crefname{table}{Tab.}{Tabs.}

\definecolor{carnelian}{rgb}{0.7, 0.11, 0.11}

\def\wacvPaperID{2587} 
\def\confName{WACV}
\def\confYear{2025}

\begin{document}

\title{Memory-efficient Continual Learning with Neural Collapse Contrastive}

\author{
Trung-Anh Dang\textsuperscript{1},
Vincent Nguyen\textsuperscript{1},
Ngoc-Son Vu\textsuperscript{2},
Christel Vrain\textsuperscript{1},\\
\textsuperscript{1}Université d'Orléans, INSA CVL, LIFO UR 4022, Orléans, France\\
\textsuperscript{2}ETIS - CY Cergy Paris University, ENSEA, CNRS, France\\
{\tt\small \{trung-anh.dang, vincent.nguyen, christel.vrain\}@univ-orleans.fr, son.vu@ensea.fr}
}


\maketitle

\begin{abstract}
    Contrastive learning has significantly improved representation quality, enhancing knowledge transfer across tasks in continual learning (CL). 
    However, catastrophic forgetting remains a key challenge, as contrastive based methods primarily focus on ``soft relationships'' or ``softness'' between samples, which shift with changing data distributions and lead to representation overlap across tasks. Recently, the newly identified Neural Collapse phenomenon has shown promise in CL by focusing on ``hard relationships'' or ``hardness'' between samples and fixed prototypes. However, this approach overlooks ``softness'', crucial for capturing intra-class variability, and this rigid focus can also pull old class representations toward current ones, increasing forgetting. Building on these insights, we propose Focal Neural Collapse Contrastive (FNC$^2$), a novel representation learning loss that effectively balances both soft and hard relationships. Additionally, we introduce the Hardness-Softness Distillation (HSD) loss to progressively preserve the knowledge gained from these relationships across tasks. Our method outperforms state-of-the-art approaches, particularly in minimizing memory reliance. Remarkably, even without the use of memory, our approach rivals rehearsal-based methods, offering a compelling solution for data privacy concerns.
\end{abstract}

\section{Introduction}
\label{sec:intro}
Unlike human learning, which is adaptive and ongoing, current deep neural networks (DNNs) are typically trained in batches, using all available data at once. To emulate human learning, these models need to support continual learning (CL), which involves acquiring new knowledge while retaining previously learned information. However, this process is hindered by the phenomenon known as ``catastrophic forgetting'' (CF) \cite{mccloskey_1989, robins1995_catastrophic_fr}, where learning new tasks with fresh data causes the model to overwrite its prior knowledge, leading to a drastic decline in performance on older tasks. To tackle this challenge, CL has gained significant attention in recent years \cite{rusu_2016, co2l, dsdm_eccv_2022, wen2024provablecontrastivecontinuallearning, shiyaoli_tcp_wacv_2024, tang_kaizen_wacv_2024, antoine_2024}. CL aims to develop methods that enable models to learn from a continuous stream of data by balancing the retention of prior knowledge with the ability to adapt to new information. Achieving this balance, known as the stability-plasticity trade-off, is crucial for preventing performance loss when integrating new tasks.

Current CL methods are predominantly based on supervised strategies, which have proven effective in preserving knowledge \cite{yoon_2018, buzzega2020darkexperiencegeneralcontinual, tiwari2022gcrgradientcoresetbased}. Notably, approaches that decouple representation learning from classifier training have shown greater resistance to forgetting compared to joint training methods \cite{co2l, fini2022cassle, madaan_2022_lump, wen2024provablecontrastivecontinuallearning}. Building on this decoupling, several supervised contrastive representation learning methods have delivered strong results \cite{co2l, wen2024provablecontrastivecontinuallearning}. However, \textcircled{1} most of these methods depend on replay buffers for storing past samples, limiting their use when data privacy is a concern. \textcircled{2} Another limitation is their reliance on inter-sample relationships, which can lead to \textit{\textbf{representation drift and overlap}} with new tasks—one of the main causes of forgetting in CL. Indeed, this issue has been highlighted in both DNNs \cite{yu_semantic_drift_2020, caccia2021_new_io_iclr} and neuroscience \cite{laura_drift_cl_neuroscience_2022}.

Neural collapse (NC), a recently discovered phenomenon characterized by highly structured and aligned neural network features, has attracted considerable attention in the deep learning community \cite{lu2021neuralcollapsecrossentropyloss, fang_pnas_2021, wenlong_ji_iclr_2022, han_nc_2021, tirer_extendeduf_2022, zhou_icml_2022}. It shows particular promise for CL by reducing representation overlap, enhancing class separation, and mitigating CF. NC achieves this by aligning feature representations with fixed prototypes, which act as optimal class-specific reference points and remain constant throughout training, significantly enhancing class separation. Additionally, prototypes can serve as class representatives, and their integration into contrastive learning reduces reliance on memory buffers. Leveraging these advantages, several CL methods \cite{yang2023neuralcollapse, yang2023neuralcollapseterminusunified, minhyuk_residual_2024} have incorporated NC. However, focusing only on sample-prototype relationships can reduce diversity, disrupt within-class data distribution, and lead to forgetting as older representations shift toward current task prototypes. To address this, we propose a supervised method using fixed, evenly spaced prototypes to enhance representation quality, minimize task overlap, and reduce memory dependence while preserving intra-class data distribution.



To summarize, the main contributions of our work are:
\begin{itemize}
    \item We reveal that ``softness'' (inter-sample relationships) and ``hardness'' (relationships with fixed prototypes) \footnote{In unsupervised contrastive-based learning, \textit{alignment} ensures similar features are mapped to similar samples, while \textit{uniformity} ensures a feature distribution that maximizes information and ideally forms a uniform hypersphere. In a supervised setting, we use the term ``softness'' to represent alignment, and introduce ``hardness'' to describe the relationships between samples and NC-based prototypes.} are critical in NC-based contrastive CL for effectively acquiring new tasks while retaining prior knowledge.
    \item In both plasticity and stability phases of learning, we introduce new loss functions that address both hardness and softness. For plasticity, we propose Focal Neural Collapse Contrastive (FNC$^2$), a loss function that combines hard and soft semantics to enhance representation learning by focusing on challenging samples. For stability, we introduce the Hardness-Softness Distillation (HSD) loss function, which preserves knowledge from both hard and soft relationships, significantly reducing forgetting.
    \item Our model surpasses state-of-the-art (SoTA) results in both replay-based and memory-free scenarios, especially excelling in settings with no stored exemplars, making it ideal for applications with strict data privacy requirements.
\end{itemize}

\section{Related Work}
\label{sec:related-work}
\subsection{Continual Learning}
\label{sec:rw-continual-learning}

CL approaches can be broadly classified into three main categories. Rehearsal-based approaches \cite{madaan_2022_lump, buzzega2020darkexperiencegeneralcontinual,co2l, wen2024provablecontrastivecontinuallearning} store a small amount of data in a memory buffer and replay them to prevent forgetting. Regularization-based methods \cite{friedemann_si_2017, sangwon_neurips_2020, fini2022cassle, dongmin, tang_kaizen_wacv_2024} penalize changes in network parameters of the current task with respect to the previous task. Meanwhile, instead of using shared parameters, the architectural-based approaches \cite{rusu_2016, yoon_2018,li_learn_tg_2019} construct task-specific parameters and allow network expansion during CL. This work focuses on regularization-based methods by devising a specific regularization loss to align the current model with the previous one. Additionally, our approach is capable of performing well in both replay-based and memory-free scenarios.

Most of the current CL methods target primarily to achieve a balance between acquiring new tasks (\textbf{plasticity}) and retaining knowledge of previous tasks (\textbf{stability}) \cite{fini2022cassle, co2l, wen2024provablecontrastivecontinuallearning, szatkowski_wacv_2023}. To attain this balance, regularization-based approaches typically employ knowledge distillation (KD) \cite{hinton_kd_2015}, which aims to transfer knowledge from the previous trained model (teacher) to the current one (student). Recent works based on KD directly minimize the divergence between their intermediate representations \cite{hou_learning_2019} or final outputs \cite{fini2022cassle, pseudo_negative_icml_2024}. Additionally, several methods have explored relational KD, which enhances knowledge retention by leveraging the sample-to-sample relationships, such as Instance-wise Relation Distillation (IRD) \cite{co2l, wen2024provablecontrastivecontinuallearning}.
Other approaches, such as \cite{asadi_prd_cl_icml_2023, li_cl_important_sampling_2024}, focus on relationships between \textbf{\textit{learnable}} class prototypes and individual samples, rather than between samples themselves. 
In this work, we utilize KD through IRD, alongside a prototype-based distillation method. Our approach differs from \cite{asadi_prd_cl_icml_2023, li_cl_important_sampling_2024} in both the use of \textbf{\textit{fixed}} rather than learnable prototypes, and in how the sample-prototype relationships are applied.

Beyond balancing plasticity and stability, recent research emphasizes the importance of \textbf{cross-task} consolidation for improving representation and reducing forgetting \cite{lode_2023, osiris_zhang_2024}. We \textbf{\textit{implicitly}} address this by designing a plasticity loss function crafted to enhance cross-task separability.

\subsection{Contrastive Learning}
\label{sec:rw-contrastive-learning}
Contrastive learning has emerged as a prominent representation learning approach, demonstrating its SoTA for different downstream tasks \cite{oord_representationlw_2018, simclr-chen20j, khosla2021supcon}. Numerous contrastive methods have been proposed and widely applied in both unsupervised \cite{oord_representationlw_2018, moco_v1_2020, chen2020_improved_bw, simclr-chen20j} and supervised settings \cite{khosla2021supcon}. 

In the context of CL, many studies \cite{fini2022cassle, scr_mai_2021, co2l, wen2024provablecontrastivecontinuallearning, li_cl_important_sampling_2024} have shown that contrastive learning is highly effective in acquiring task-invariant representations, which significantly mitigates the primary cause of forgetting — data imbalances between previous and current tasks. Among these, Co$^2$L \cite{co2l} is the first method to apply supervised contrastive learning in CL. Subsequently, CILA \cite{wen2024provablecontrastivecontinuallearning} improved Co$^2$L by analyzing the importance of coefficients for the distillation loss. Recently, CCLIS \cite{li_cl_important_sampling_2024} emerged as a SoTA method by preserving knowledge through importance sampling to recover previous data distributions. However, unlike Co$^2$L and CILA, CCLIS cannot operate without a memory buffer, which limits its efficiency in many real-world applications where data privacy is a concern.

\begin{figure*}[t]
  \centering
   \includegraphics[width=0.8\linewidth]{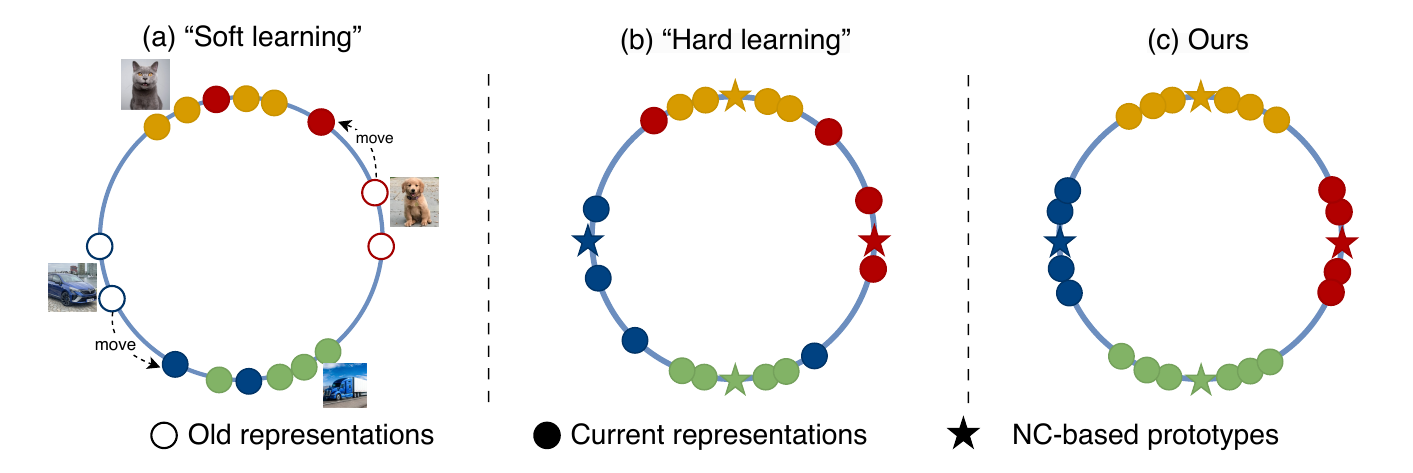}

   \caption{Our method mitigates the drawbacks of both ``soft'' and ``hard'' learning by using fixed, equidistant prototypes to minimize cluster overlap. (a) Overlap between new and old classes happens due to non-fixed class clusters and representation drift. (b) Strong alignment with prototypes can overly cluster current task representations and pull older ones into new clusters, especially in CL with few old samples. Moreover, mixed-feature samples should be placed between classes, not tightly aligned with prototypes. (c) Our method considers both inter-sample and sample-prototype relationships, maintaining distinct cluster representations and preserving distribution within each cluster.}
   \label{fig:comparasion-ours}
\end{figure*}

Contrastive learning generates augmented views of each sample, bringing positive pairs closer and pushing negative pairs apart, promoting representation invariance to augmentations. Most contrastive methods focus on learning representations through relationships between samples, referred to as soft relationships or ``softness''. In CL, relying solely on softness preserves class diversity and data distribution but can cause task representation overlap, as shown in \cref{fig:comparasion-ours}a. In this paper, we propose a novel supervised contrastive loss for CL that reduces memory dependence and resolves this overlap while maintaining class distribution.

\subsection{Neural Collapse (NC)}
\label{sec:rw-neural-collapse}
A recent study \cite{papyan_2020} identified neural collapse (NC) phenomenon, where, at the end of training on a balanced dataset, class features collapse to their class means, aligned with a simplex equiangular tight frame (ETF). This finding has led to further research showing that NC represents global optimality in balanced training with cross-entropy \cite{lu2021neuralcollapsecrossentropyloss, florian_icml_2021, fang_pnas_2021, zhu_neurips_2021, wenlong_ji_iclr_2022} and mean squared error \cite{poggio_explicit_ra_2020, han_nc_2021, zhou_icml_2022, tirer_extendeduf_2022} loss functions. Inspired by NC, studies such as \cite{larue_nicolas_seeable_2022} have used fixed simplex ETF points with modified contrastive loss to achieve NC, while \cite{yang_induce_nc_2022} induced NC under imbalanced data conditions by fixing the classifier.


\textbf{Inducing NC for CL}. Building on \cite{galanti_ontr_2022}, which showed that NC persists when transferring models to new samples or classes, several CL studies have leveraged NC to mitigate forgetting \cite{yang2023neuralcollapse, yang2023neuralcollapseterminusunified}. These works pre-assign a group of classifier prototypes as a simplex ETF for all tasks and then align sample representations to their corresponding prototypes. The relationship between samples and prototypes, which this learning approach focuses on, is referred to as hard relationships, or ``hardness''. For instance, \cite{yang2023neuralcollapse, yang2023neuralcollapseterminusunified} employed the dot-regression (DR) loss proposed in \cite{yang_induce_nc_2022} for NC-based CL, which is considered a hard learning method. Since the prototype group remains consistent, these methods prevent task overlap but risk reducing diversity and disrupting within-class distribution, potentially leading to forgetting. Our method addresses this issue by integrating NC directly into the loss function, combining both softness and hardness to preserve data distribution and avoid task overlap in CL.

The concept of NC can be presented as follows.

\textbf{Definition 1.} A simplex Equiangular Tight Frame (ETF) is a collection of K vectors: $\mathbf{Q} = \{\mathbf{q}_k\}_{k=1}^{K}$, each vector $\mathbf{q}_k \in \mathbb{R}^{d}$, $K \le d+1$, which satisfies:

\begin{equation}
    \mathbf{Q} = \sqrt{\frac{K}{K-1}}\mathbf{U}\left(\mathbf{I}_K-\frac{1}{K}\mathbf{1}_K\mathbf{1}_K^T\right),
    \label{eq:etf}
\end{equation}
where $\mathbf{U} \in \mathbb{R}^{d \times K}$ is an orthogonal basis and $\mathbf{U}^{T}\mathbf{U}=\mathbf{I_K}$, $\mathbf{I}_K$ is an identity matrix and $\mathbf{1}_K$ is an all-ones vector.

Each vector $\mathbf{q}_k$ has the same $\ell_2$ norm, and any two distinct vectors consistently produce an inner product of $-\frac{1}{K-1}$, which is the lowest possible cosine similarity for K equiangular vectors in $\mathbb{R}^d$. This geometric relationship can be described as
\begin{equation}
    \mathbf{q}^T_{i}\mathbf{q}_{j}=\frac{K}{K-1}\delta_{i,j}-\frac{1}{K-1},\ \ \forall i, j\in[1,K],
    \label{eq:cosine-similarity-etf}
\end{equation}
where $\delta_{i, j}=1$ in case of $i=j$, and 0 otherwise.

After that, the NC phenomenon can be formally characterized by the following four attributes~\cite{papyan_2020}:

NC1: Features from the last layer within the same class converge to their intra-class mean, such that the covariance $\Sigma^{k}_V\rightarrow\mathbf{0}$. Here, $\Sigma^{k}_V=\mathrm{Avg}_{i}\{(\boldsymbol{\nu}_{k,i}-\boldsymbol{\mu}_{k})(\boldsymbol{\nu}_{k,i}-\boldsymbol{\mu}_{k})^T\}$, where $\boldsymbol{\nu}_{k,i}$ is the feature of sample $i$ in class $k$, and $\boldsymbol{\mu}_{k}$ is the intra-class feature mean.

NC2: After centering by the global mean, intra-class means align with simplex ETF vertices, i.e., $\{\tilde{\boldsymbol{\mu}}_k\}$, $1 \le k \le K$ satisfy \cref{eq:cosine-similarity-etf}, where $\tilde{\boldsymbol{\mu}}_k=(\boldsymbol{\mu}_k - \boldsymbol{\mu}_G) / \|  \boldsymbol{\mu}_k - \boldsymbol{\mu}_G \|$ and global mean $\boldsymbol{\mu}_G=\frac{1}{K}{\sum_{k=1}^{K}{\boldsymbol{\mu}_k}}$;

NC3:  Intra-class means centered by the global mean align with their classifier weights, leading to the same simplex ETF, i.e., $\tilde{\boldsymbol{\mu}}_k=\boldsymbol{w}_k/ \| \boldsymbol{w}_k \|$, where $1\le k \le K$ and $\boldsymbol{w}_k$ is the classifier weight of class $k$;

NC4: When NC1-NC3 hold, model predictions simplify to selecting the nearest class center, represented as $\text{argmax}_k\langle\mathbf{z}, \boldsymbol{w}_k\rangle=\text{argmin}_k||\mathbf{z}-\boldsymbol{\mu}_k||$, where $\langle \cdot , \cdot \rangle$ denotes the inner product operator, and $\mathbf{z}$ is the output of the model.






\section{Preliminaries}
\label{sec:preliminaries}

\subsection{Problem Setup}
\label{sec:problem-setup}

In the general supervised CL scenario, we have a sequence of training datasets, which is drawn from non-stationary data distributions for each task. Namely, let $t$ be the task index, where $t \in \{1,...,T\}$, and $T$ represents the maximum number of tasks. The dataset for the $t$-th task, denoted by $\mathcal{D}_t$, consists of $N_t$ supervised pairs: $\mathcal{D}_t=\{(\mathbf{x}_i,y_i)\}_{i=1}^{N_t}$, along with the set of classes $\mathcal{C}_t$. CL comprises a variety of scenarios; however, in this work, we focus specifically on two popular settings: class-incremental learning (Class-IL) and task-incremental learning (Task-IL). In both settings, there is no overlap in class labels across tasks, ensuring $\mathcal{C}_t \cap \mathcal{C}_{t'} = \emptyset$ for any two distinct tasks $t \neq t'$. For Task-IL, the learned model additionally has access to the task label during the testing phase.

\subsection{Supervised Contrastive Learning}
This section details the SupCon algorithm \cite{khosla2021supcon}, which inspires the ``softness'' component of our approach. Suppose that in each batch $B$ of $N$ samples, SupCon firstly creates two randomly augmented versions of each sample in the batch, making each batch now contain $2N$ views: $\left|B\right|=2N$. After that, given the feature extractor $f$, each view $\mathbf{x}_i$ in the batch is mapped into a unit d-dimensional Euclidean sphere through a linear projector $g$ as $\mathbf{z}_i=h(\mathbf{x}_i)$, where $h = g \circ f$. Consequently, generic presentations are learned through the minimization of the following loss:
\begin{equation}
    \mathcal{L}_{SupCon}=\sum_{i=1}^{2N}\frac{-1}{\left|P(i)\right|}\sum_{j \in P(i)}\log(\frac{e^{\langle\mathbf{z}_i \cdot \mathbf{z}_j \rangle/\tau}}{\sum_{k \in A(i)}e^{\langle\mathbf{z}_i \cdot \mathbf{z}_k\rangle/\tau}})
    \label{eq:sup-con}
\end{equation}

where $\langle\cdot\rangle$ is the cosine similarity, $\tau > 0$ is the temperature factor, $A(i) = \{1..2N\} \setminus \{i\}$, and $P(i)$ is the index set of positive views with the anchor $\mathbf{x}_i$, denoted as:

\begin{equation}
    \label{eq:set-postive-index}
    P(i) = \{p \in \{1...2N\} | y_p=y_i, p \neq i\}    
\end{equation}

\noindent\textbf{Focal contrastive learning.} 
Despite their advantages, contrastive learning methods often struggle to reduce intra-class feature dispersion. They rely heavily on positive/negative pairs, but most samples are easy to contrast, resulting in minimal loss gradients and scattered intra-class samples. To address this issue, inspired by focal loss~\cite{lin2018focallossdenseobject}, ~\cite{focalcontrastive} introduced the focal contrastive loss. This approach emphasizes hard positive views—those with low cosine similarity to the anchor and thus lower prediction probability. Since hard positives are more influential in contrastive loss, they lead to clearer class clustering. Building on these insights, we propose focal contrastive loss in the context of NC for CL, as detailed in \cref{sec:fnc3}.


\section{Methodology}
\subsection{Motivation}
\label{sec:motivation}
Our method uses a two-stage learning process, as in ~\cite{co2l}. First, we learn a representation, which is then used to train the classifier. The main objective of CL is to balance two goals: learning new tasks (plasticity) and preserving the knowledge from previous tasks (stability). The overall loss can be described as:

\begin{equation}
    \label{eq:general-cl}
    \mathcal{L}_{overall}=\mathcal{L}_{plasticity} + \mathcal{L}_{stability}
\end{equation}

\textbf{\textit{Remark.}} While plasticity and stability are central to most CL algorithms, some very recent works have highlighted the importance of cross-task consolidation \cite{lode_2023, osiris_zhang_2024}. Although our approach does not explicitly incorporate a cross-task consolidation term, we will show that it is implicitly accounted for in our $\mathcal{L}_{plasticity}$.

\noindent \textbf{a. Plasticity.} Previous contrastive learning approaches, like SupCon \cite{khosla2021supcon}, use sample relationships but allow class representations to shift, as shown in \cref{fig:comparasion-ours}a, causing overlap with current clusters and leading to forgetting. We term these methods ``soft plasticity''. To address this issue, inspired by Neural Collapse, recent works \cite{yang2023neuralcollapse, yang2023neuralcollapseterminusunified} propose using fixed, equidistant prototypes as the optimal class means at the end of training, as shown in \cref{fig:comparasion-ours}b. These methods, which we term ``hard plasticity'', focus on aligning sample representations with their assigned prototypes.


While these ``hard'' methods tightly align representations with prototypes, they have drawbacks. First, they neglect sample relationships, leading to representations clustered only around prototypes, which is not ideal since some samples share characteristics with multiple classes and should lie between them. Second, tightly aligning samples with current prototypes can pull old class representations towards the current task’s prototypes, causing forgetting, as shown in \cref{fig:comparasion-ours}b. To harness the benefits of NC while avoiding the pitfalls of hard learning, we integrate both soft and hard learning. This approach maintains cluster distribution within each class and preserves separation from clusters of both current and past classes, as shown in \cref{fig:comparasion-ours}c.

\begin{figure}[t]
  \centering
   \includegraphics[width=0.86\linewidth]{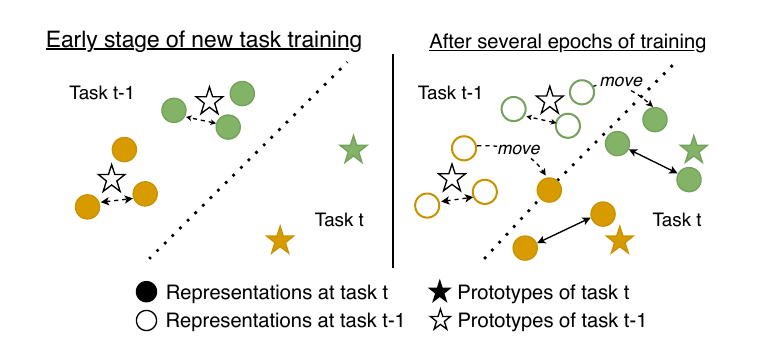}

   \caption{Shifts in representations of current samples at task $t$ between the beginning of training and after several epochs.}
   \label{fig:comare-ird-sprd}
\end{figure}

\noindent  \textbf{b. Stability.} Co$^2$L, the pioneering work in continual contrastive learning, employs IRD as 
$\mathcal{L}_{stability}$.
The ``soft stability’’ IRD is designed to preserve the relationships between samples in the old and new feature spaces. However, as training progresses, its effectiveness diminishes, as demonstrated in \cref{fig:comare-ird-sprd}: 
\begin{enumerate}
    \item At the start of training task $t$, the representations of all current samples match those from task $t-1$ because no training has occurred yet. Initially, sample representations are consistent across tasks, but as training progresses, they gradually shift towards their respective class clusters.
    \item Over many epochs, the increasing discrepancy between the current and past feature spaces leads to a rise in IRD loss. This shift occurs because representations approach optimal positions, reducing the significance of 
$\mathcal{L}_{stability}$ based on IRD to the overall loss 
$\mathcal{L}_{overall}$ over time.
\end{enumerate} 
To address this limitation, we introduce a new distillation loss, termed ``hard stability’’, inspired by NC phenomenon, along with a novel strategy to enhance IRD's effectiveness throughout the training process. Both of these will be detailed in \cref{sec:hsd}. As presented in \cref{fig:comparasion-ours}, both soft and hard learning methods suffer from representation drift, especially in settings with limited or no memory. Our approach effectively mitigates this issue. However, as the memory capacity increases, the impact of representation drift in soft and hard learning methods diminishes. Consequently, our method demonstrates a significant advantage in low-memory regimes, where the problem of representation drift is more pronounced.

\subsection{Overview of the proposed method}
\begin{figure*}[t]
  \centering
   \includegraphics[width=0.9\linewidth]{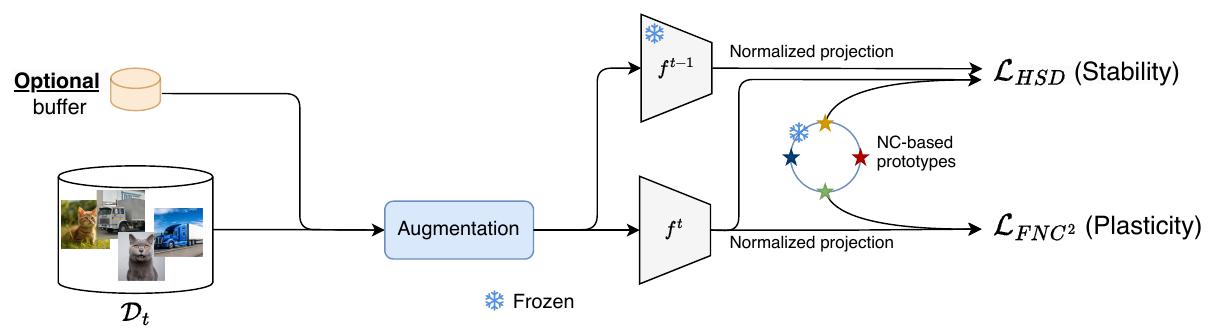}

   \caption{\textbf{Overall architecture of our method.} Augmented samples from each batch are fed into the current model $f^t$ to learn new knowledge via $\mathcal{L}_{FNC^2}$ and the frozen previous model $f^{t-1}$ for distillation using $\mathcal{L}_{HSD}$. The buffer is optional, and NC-based prototypes are directly involved in both loss functions during training. }
   \label{fig:overall-methods}
\end{figure*}

\cref{fig:overall-methods} provides an overview of our method (see \cref{sec:problem-setup} for setting details). 
In our approach, memory is optional, and when used, we employ the Reservoir sampling strategy \cite{reservoir_sampling_1985} to fill a fixed-size buffer $\mathcal{M}$. After each task beyond the first, current samples are combined with buffered samples, and each sample is drawn independently with equal probability for the mini-batch. We first predefine a set of fixed equidistant prototypes as the vertices of an ETF. 
We denote this prototype set as $\mathbf{P} = \{\mathbf{p_i}\}_{i=1}^{K}$, $K$ is the number of prototypes, corresponding to the number of classes. These prototypes are used as equidistant optimal points in the feature space. We utilize these prototype points in both learning new tasks and distilling old tasks through their direct use in the corresponding loss functions. 

The overall objective of our method is:
\begin{equation}
    \label{eq:final-objective}
    \mathcal{L} = \mathcal{L}_{FNC^2} + \mathcal{L}_{HSD}
\end{equation}
with details on each new loss function provided in the following sections.

\subsection{Hardness-Softness Plasticity}
\label{sec:fnc3}
Inspired by NC phenomenon \cite{papyan_2020} and the concept of focal loss for addressing batch imbalance during training, we introduce a novel loss called Focal Neural Collapse Contrastive (FNC$^2$), defined as:
\begin{align}
    \mathcal{L}_{FNC^2} = -\sum_{i=1}^{2N} \frac{1}{|P(i)| + 1} \Bigg(& \sum_{\mathbf{z}_{j} \in P(i)} (1-c_{ij})^{\gamma} \log(c_{ij}) \nonumber \\
    &+ (1-r_{i})^{\gamma} \log(r_{i}) \Bigg)
    \label{eq:fnc3-loss}
\end{align}

where
\begin{equation}
    c_{ij}=\frac{e^{\langle\mathbf{z}_i \cdot \mathbf{z}_j\rangle/\tau}}{\sum_{k \neq i}{e^{\langle\mathbf{z}_i \cdot \mathbf{z}_k\rangle/\tau}} + \underbrace{\sum_{\mathbf{p}_l \in \mathbf{P}_{1:t-1}}{e^{\langle{\mathbf{z}_i \cdot \mathbf{p}_l}\rangle/\tau}}}_{\text{cross-task}}}
    \label{eq:c-ij}
\end{equation}
\begin{equation}
    r_{i}=\frac{e^{\langle\mathbf{z}_i \cdot \mathbf{p}_{\mathbf{z}_i}\rangle/\tau}}{\sum_{k \neq i}{e^{\langle\mathbf{z}_i \cdot \mathbf{z}_k\rangle/\tau}} + \sum_{\mathbf{p}_l \in \mathbf{P}_{1:t-1}}{e^{\langle\mathbf{z}_i \cdot \mathbf{p}_l\rangle/\tau}}}
    \label{eq:r-ij}
\end{equation}

Here, $P(i)$ is the set of positive indexes for each anchor, as defined in \cref{eq:set-postive-index}, $|P(i)|$ represents its cardinality, and $\gamma$ is the focusing hyperparameter. Additionally, $\mathbf{p}_{\mathbf{z}_i}$ is the prototype (specifically, the ETF vertex corresponding to label $y_i$), and $\mathbf{P}_{1:t-1}$ represents the set of prototypes used in all previous tasks.

Intuitively, for each sample in the current task, we pull positive samples closer to the anchor and push negative samples away, forming clusters that are then pulled towards their optimal prototypes, as shown in \cref{eq:c-ij} and \cref{eq:r-ij}. This approach helps the model learn both hardness and softness information. By incorporating $(1-c_{ij})^{\gamma}$ and $(1-r_{i})^{\gamma}$, $\mathcal{L}_{FNC^2}$ emphasizes hard samples - those that are positive but far from the anchor or far from their prototype. These hard samples are crucial in contrastive learning, as they significantly affect intra-class sample distribution, unlike easy-to-contrast samples. A larger $\gamma$ further focuses the model on training from samples that are distant from their positive views and prototypes. To reduce reliance on memory, we use prototypes from previous tasks as representative points for past samples and include their cosine similarity with the anchor in $c_{ij}$ and $r_{i}$. This allows our method to use old prototypes as negative points, similar to \textit{\textbf{cross-task consolidation}} as described in \cite{lode_2023, osiris_zhang_2024}, aiding in distinguishing between current and old classes. This approach effectively mimics stored samples and facilitates robust learning even without past task samples, serving as a form of \textbf{\textit{pseudo-replay}} where old prototypes are effectively reintroduced. We will present performance in both memory-based and memory-free settings in \cref{sec:results}.

When using memory, we employ an asymmetric version of this loss, where only current samples serve as anchors and buffer samples are used solely as negative points.

\subsection{Hardness-Softness Distillation}
\label{sec:hsd}
Our new distillation loss, which capitalizes on both hardness and softness plasticity during training, is described as:
\begin{align}
    \mathcal{L}_{HSD} &= (1 - \alpha) \underbrace{\sum_{i=1}^{2N}{-\mathbf{o}^{t-1}(\mathbf{x}_i) \cdot \log(\mathbf{o}^{t}(\mathbf{x}_i))}}_{\mathcal{L}_{IRD}}
    \nonumber \\
    &\quad + \alpha \underbrace{\sum_{i=1}^{2N} -\textbf{q}^{t-1}(\mathbf{x}_i;\mathbf{P}_{1:t}) 
\cdot\log(\textbf{q}^t(\mathbf{x}_i;\mathbf{P}_{1:t}))}_{\mathcal{L}_{S-PRD}}
    \label{eq:hsd}
\end{align}
with $\alpha = max(0, \frac{e - e_0}{E})$, $e$ is the epoch index, $e_0$ is the number of epochs used for the warm-up period, and $E$ is the number of epochs (details are provided in the Appendix). 
Here, $\mathcal{L}_{IRD}$ (used in \cite{co2l}) represents ``soft stability'' while the proposed Sample-Prototype Relation Distillation loss $\mathcal{L}_{{S-PRD}}$ embodies ``hard stability''.

In $\mathcal{L}_{IRD}$, $\mathbf{o}^t(\mathbf{x}_i)=[o^t_{i,1},...,o^t_{i,i-1},o^t_{i},o^t_{i,i+1},...,o^t_{i,2N}]$, 
each $o^t_{i,j}$ is: $o_{i,j}^t=\frac{e^{\langle\mathbf{z}_i^t \cdot \mathbf{z}_j^t\rangle/\kappa}}{\sum_{k\neq{i}}e^{\langle\mathbf{z}_i^t \cdot \mathbf{z}_k^t\rangle/\kappa}}$,
where $t \ge 1$ is the task index and $\kappa$ is the temperature hyperparameter.
In $\mathcal{L}_{{S-PRD}}$, $\mathbf{P}_{1:t} = \{\mathbf{p}_s\}_{s=1}^{S}$ is the set of prototypes used from task $1$ to task $t$.
Besides, we have $\textbf{q}^t{(\mathbf{x}_i;\mathbf{P}_{1:t})}=[q^t_{i,1}, q^t_{i,2},...,q^t_{i,S}]$, and each ${q}_{i,j}^t$ is computed as: $q_{i,j}^t=\frac{e^{\langle\mathbf{z}_i^t \cdot \mathbf{p}_j\rangle/\zeta}}{\sum_{s=1}^{S}{e^{\langle\mathbf{z}_i^t \cdot \mathbf{p}_s\rangle/\zeta}}}$, given the temperature factor $\zeta$.

$\mathcal{L}_{S-PRD}$ preserves hardness by maintaining sample-prototype relationships, while $\mathcal{L}_{IRD}$ focuses on softness, capturing the evolving relationships between samples during learning. While $\mathcal{L}_{S-PRD}$ strongly preserves old knowledge by regularizing sample-prototype relationships, it may hinder learning if used exclusively, as it restricts the adaptation of representations. Early in training, $\mathcal{L}_{IRD}$ is crucial as it supports learning plasticity, while $\mathcal{L}_{S-PRD}$ is less effective. As training progresses and representations approach their prototypes, $\mathcal{L}_{S-PRD}$ becomes more beneficial, enhancing distillation.
That is the motivation behind the idea of combining $\mathcal{L}_{IRD}$ and $\mathcal{L}_{S-PRD}$ in $\mathcal{L}_{HSD}$ (Hardness-Softness Distillation) with weights that shift over time. After an initial warm-up period ($e_0$ epochs), we gradually decrease $\mathcal{L}_{IRD}$ and increase $\mathcal{L}_{S-PRD}$ to balance learning and information preservation effectively.

\section{Experiments}
\label{sec:experiments}

\subsection{Datasets and Implementation details}
\label{sec:datasets}
\noindent \textbf{Datasets.} As in other continual contrastive learning works, we use thee datasets: Seq-Cifar-10, Seq-Cifar-100, and Seq-Tiny-ImageNet for all experiments. We consider the class-IL and task-IL settings, described in \cref{sec:problem-setup}. Seq-Cifar-10 is created from Cifar-10 \cite{krizhevsky2009learning} and divided into five tasks with two classes for each task, Seq-Cifar-100 contains five tasks (20 classes/task) built from Cifar-100 \cite{krizhevsky2009learning}, and Seq-Tiny-ImageNet has 10 tasks (20 classes/task) built from Tiny-ImageNet \cite{le2015tinyiv}.

\noindent \textbf{Implementation details.} We train our method using ResNet-18 \cite{resnet18} as the backbone, and we remove the last layer of the backbone as \cite{co2l, fini2022cassle}. Similar to previous works \cite{simclr-chen20j, zbontar2021barlowtwinsselfsupervisedlearning, co2l}, we add a two-layer projection MLP on top of the backbone, followed by ReLU activation functions to map the output of the backbone into $d$-dimension embedding space, where $d=128$ with Seq-Cifar-10 and Seq-Cifar-100, $d=256$ with Seq-Tiny-ImageNet. With a batch size of 512, we train the backbone with 500 epochs for the initial task and 100 epochs for other tasks as in \cite{co2l} for all datasets. Besides, we run with buffer sizes 0, 200 and 500 to evaluate the performance of the model with different memory settings. Regarding the value of $\gamma$ in $\mathcal{L}_{FNC^2}$, we choose $\gamma=1$ for Seq-Cifar-10, and $\gamma=4$ for all other datasets. Details about the selection of $\gamma$ and other hyperparameters are provided in the Appendix. Since Co$^2$L \cite{co2l} does not report results on Seq-Cifar-100 and results in the memory-free setting on Seq-Tiny-ImageNet, we run Co$^2$L on these cases for comparison. For evaluation, we train a classifier on top of the learned backbone using last-task samples and buffered samples with 100 epochs.

\noindent \textbf{Evaluation metrics.} Similar to \cite{co2l, wen2024provablecontrastivecontinuallearning}, we evaluate the quality of the learned encoder $f^t$ by training a classifier $h^t$ on top of the frozen encoder $f^t$, using only the current training dataset $\mathcal{D}_t$ and the samples from the memory $\mathcal{M}$.
As defined in \cite{chaudhry_2018} as well as in other CL methods \cite{co2l, fini2022cassle, wen2024provablecontrastivecontinuallearning}, we compute the Average Accuracy (AA) on the test dataset and all accuracies $A_{T, k}$ of each task $k$ after learning the final task $T$. The equation for AA is defined as: 
\begin{equation}
    \label{eq:average-accuracy}
    AA = \frac{1}{T}\sum_{k=1}^{T}A_{T,k}
\end{equation}

\noindent\textbf{Considered methods.} To focus on learning ability with limited or no memory, we compare our method only with recent approaches that can operate in both settings. Since CCLIS \cite{li_cl_important_sampling_2024} and LODE \cite{lode_2023} cannot operate in memory-free scenarios, their results are excluded in \cref{tab:acc_result}. As shown in \cref{tab:memory-free-ability}, only Co$^2$L \cite{co2l}, the recent method CILA \cite{wen2024provablecontrastivecontinuallearning}, and ours can run without a memory buffer. Additionally, we compare our results with other well-known supervised methods, including ER \cite{riemer2019learninglearnforgettingmaximizing}, iCaRL \cite{rebuffi2017icarl}, GEM \cite{lopezpaz2017gem}, GSS \cite{aljundi2019gradientbasedsampleselection}, DER \cite{buzzega2020darkexperiencegeneralcontinual}, Co$^2$L \cite{co2l}, GCR \cite{tiwari2022gcrgradientcoresetbased}, and CILA \cite{wen2024provablecontrastivecontinuallearning}.

\subsection{Main Results}
\label{sec:results}
\cref{tab:acc_result} reports results for buffer sizes 0 and 200 to highlight memory-free and limited memory scenarios. Results for a buffer size of 500 are provided in the Appendix. Although CILA \cite{wen2024provablecontrastivecontinuallearning} can operate without memory, results for this setting are not available due to the lack of publicly accessible code, so we cannot include them. 

As shown in \cref{tab:acc_result}, our model significantly outperforms recent replay-based methods across various CL settings, datasets, and buffer sizes. Notably, without using a buffer, our method outperforms Co$^2$L in all settings and datasets. For Seq-Cifar-10, our method exceeds Co$^2$L by approximately $10\%$ in Class-IL and $7.76\%$ in Task-IL. Similarly, for Seq-Cifar-100, our method outperforms Co$^2$L by $5.68\%$ for Class-IL and by $5.96\%$ for Task-IL. On Seq-Tiny-ImageNet, our method surpasses Co$^2$L in both Class-IL and Task-IL by $1.11\%$ and $3.6\%$, respectively. This improvement is less pronounced than on the CIFAR-10/100 datasets because the current evaluation protocol trains the classifier with samples from the current task and only a limited number from each old class. Consequently, the less significant improvements may stem from the classifier rather than the representation.

When compared to other methods with a memory buffer size of 200, our method outperforms all except for Seq-Cifar-100 in the Task-IL setting, where it lags behind GCR \cite{tiwari2022gcrgradientcoresetbased}. Notably, in the memory-free scenario, our method surpasses all replay methods with a buffer size of 200 on Seq-Cifar-10 and nearly matches the top result on Seq-Tiny-ImageNet, achieved by CILA. Although memory-free results for CILA are not available, our method outperforms their buffer size 200 results by 2.2\% on Seq-Cifar-10, highlighting its robustness in both limited memory and memory-free settings. With SoTA performance in memory-free scenarios, our model is well-suited for real-world applications where data storage is a concern.

Additionally, the Appendix includes average forgetting results to assess how well our method retains knowledge of previous tasks compared to other baselines.

\begin{table}
  \centering

    \resizebox*{!}{0.24\columnwidth}{
        
\begin{tabular}{@{}l|cc@{}}
    \toprule
    \textbf{Type} & \textbf{Method} \\
    \midrule
    \textbf{Memory-based} & ER \cite{riemer2019learninglearnforgettingmaximizing}, iCaRL \cite{rebuffi2017icarl}, GEM \cite{lopezpaz2017gem}, GSS \cite{aljundi2019gradientbasedsampleselection}, \\
    & DER \cite{buzzega2020darkexperiencegeneralcontinual}, \textbf{Co$^2$L} \cite{co2l}, GCR \cite{tiwari2022gcrgradientcoresetbased}, LODE \cite{lode_2023}, \\
    & \textbf{CILA} \cite{wen2024provablecontrastivecontinuallearning}, CCLIS \cite{li_cl_important_sampling_2024}, \textbf{Ours} \\
    \midrule
    \textbf{Memory-free ability} & \textbf{Co$^2$L} \cite{co2l}, \textbf{CILA} \cite{wen2024provablecontrastivecontinuallearning}, \textbf{Ours} \\
    \bottomrule
\end{tabular}

    }
  \caption{Among supervised CL, ours is one of the few that operates without memory.}
  \label{tab:memory-free-ability}
\end{table}

\begin{table*}[!t]
    \centering
    \resizebox*{!}{0.6\columnwidth}{
        \begin{tabular}{clccccccc}
\hline

\multirow{2}{*}{\textbf{Buffer}}&\textbf{Dataset}&\multicolumn{2}{c}{\textbf{Seq-Cifar-10}}&\multicolumn{2}{c}{\textbf{Seq-Cifar-100}}&\multicolumn{2}{c}{\textbf{Seq-Tiny-ImageNet}}\\
            &\textbf{Scenario}&\textbf{Class-IL}&\textbf{Task-IL}&\textbf{Class-IL}&\textbf{Task-IL}&\textbf{Class-IL}&\textbf{Task-IL}\\
\hline
    \multirow{2}{*}{0}
    &Co$^2$L \cite{co2l}&58.89$\pm$2.61&86.65$\pm$1.05&26.89$\pm$0.78&51.91$\pm$0.63&13.43$\pm$0.57&40.21$\pm$0.68\\

    &\text{Ours}&\textbf{69.26$\pm$0.32}&\textbf{94.41$\pm$0.43}&\textbf{32.57$\pm$0.55}&\textbf{57.87$\pm$0.62}&\textbf{14.54$\pm$0.52}&\textbf{43.81$\pm$0.47}\\
    
\hline
    \multirow{9}{*}{200}
    &ER \cite{riemer2019learninglearnforgettingmaximizing}&44.79$\pm$1.86&91.19$\pm$0.94&21.78$\pm$0.48&60.19$\pm$1.01&8.49$\pm$0.16&38.17$\pm$2.00\\
    &iCaRL \cite{rebuffi2017icarl}&49.02$\pm$3.20&88.99$\pm$2.13&28.00$\pm$0.91&51.43$\pm$1.47&7.53$\pm$0.79&28.19$\pm$1.47\\
    &GEM \cite{lopezpaz2017gem}&25.54$\pm$0.76&90.44$\pm$0.94&20.75$\pm$0.66&58.84$\pm$1.00&-&-\\
    &GSS \cite{aljundi2019gradientbasedsampleselection}&39.07$\pm$5.59&88.80$\pm$2.89&19.42$\pm$0.29&55.38$\pm$1.34&-&-\\
    &DER \cite{buzzega2020darkexperiencegeneralcontinual}&61.93$\pm$1.79&91.40$\pm$0.92&31.23$\pm$1.38&63.09$\pm$1.09&11.87$\pm$0.78&40.22$\pm$0.67\\
    &Co$^2$L \cite{co2l}&65.57$\pm$1.37&93.43$\pm$0.78&27.38$\pm$0.85&53.94$\pm$0.76&13.88$\pm$0.40&42.37$\pm$0.74\\
    &GCR \cite{tiwari2022gcrgradientcoresetbased}&64.84$\pm$1.63&90.80$\pm$1.05&33.69$\pm$1.40&\textbf{64.24$\pm$0.83}&13.05$\pm$0.91&42.11$\pm$1.01\\
    &CILA \cite{wen2024provablecontrastivecontinuallearning}&67.06$\pm$1.59&94.29$\pm$0.24&-&-&14.55$\pm$0.39&44.15$\pm$0.70\\
    &\text{Ours}&\textbf{72.63$\pm$0.78}&\textbf{95.31$\pm$0.32}&\textbf{34.04$\pm$0.42}&{59.46$\pm$0.65}&\textbf{15.52$\pm$0.53}&\textbf{44.59$\pm$0.72}\\
\hline

\end{tabular}
    }
    \small
    \caption{Results for our method compared with other supervised baselines, with memory sizes 0 and 200, are averaged over five trials (best results in each column are highlighted in bold).}
    \label{tab:acc_result}
\end{table*}

\subsection{Ablation Studies}
\label{sec:ablation-studies}
\noindent \textbf{Effectiveness of FNC$^2$.} To evaluate the efficiency of learning plasticity via $\mathcal{L}_{FNC^2}$, we compare it with the asymmetric version of SupCon loss ($\mathcal{L}_{SupCon}^{asym}$) \cite{co2l}, in combination with different using distillation methods scenario in the Class-IL setting. Note that with buffer size 0, the asymmetric loss $\mathcal{L}_{SupCon}^{asym}$ become the original SupCon loss ($\mathcal{L}_{SupCon}$). The results in \cref{tab:plasticity-verfication} show that when stability loss is absent, the performance with both $\mathcal{L}_{FNC^2}$ and $\mathcal{L}_{SupCon}^{asym}$ drops significantly, with both yielding approximately the same results. When using distillation methods, $\mathcal{L}_{FNC^2}$ outperforms $\mathcal{L}_{SupCon}^{asym}$ in all cases, especially in the case of no memory or small memory size (200).

\begin{table}
  \centering

    \resizebox*{!}{0.33\columnwidth}{
        \begin{tabular}{lccccc}
    \toprule
    \multirow{2}{*}{\textbf{Plasticity}} & \multirow{2}{*}{\textbf{Stability}} & \multicolumn{3}{c}{\textbf{Buffer size}}\\
    &&0&200&500 \\
    \midrule
    $\mathcal{L}_{FNC^2}$ & \ding{55}&53.59$\pm$0.63&53.62$\pm$0.81&58.71$\pm$0.93\\
    $\mathcal{L}_{SupCon}^{asym}$ & \ding{55}&53.25$\pm$1.70&53.57$\pm$1.03&58.56$\pm$0.85\\
    $\mathcal{L}_{SupCon}^{asym}$ & $\mathcal{L}_{IRD}$&58.89$\pm$2.61&65.57$\pm$1.37&74.26$\pm$0.77\\
    $\mathcal{L}_{FNC^2}$ & $\mathcal{L}_{IRD}$&63.65$\pm$0.55&70.54$\pm$0.95&74.81$\pm$1.12\\
    $\mathcal{L}_{FNC^2}$ & $\mathcal{L}_{HSD}$&\textbf{69.26$\pm$0.32}&\textbf{72.63$\pm$0.78}&\textbf{75.51$\pm$0.52}\\
    \bottomrule
\end{tabular}
    }
    \caption{Performance comparison of $\mathcal{L}_{FNC^2}$ with $\mathcal{L}_{SupCon}^{asym}$ in Class-IL setting on Seq-Cifar-10. The test $\mathcal{L}_{SupCon}^{asym}$ with $\mathcal{L}_{S-PRD}$ is omitted due to incompatibility, and in case of buffer size 0, $\mathcal{L}_{SupCon}^{asym}$ become $\mathcal{L}_{SupCon}$.}
    \label{tab:plasticity-verfication}
\end{table}

\noindent \textbf{Effectiveness of HSD.} In assessing the ability of $\mathcal{L}_{HSD}$ to preserve prior knowledge, we compare cases without and with different distillation methods, incorporating the plasticity loss $\mathcal{L}_{FNC^2}$. The results in \cref{tab:stability-verification} demonstrate that using either $\mathcal{L}_{IRD}$ \cite{co2l} or $\mathcal{L}_{S-PRD}$ individually yields minimal changes in performance. However, when both methods are combined in $\mathcal{L}_{HSD}$, performance consistently improves across all buffer sizes.

\begin{table}
  \centering

  \resizebox*{!}{0.47\columnwidth}{
    \begin{tabular}{lccc}
    \toprule
    \textbf{Description} & \textbf{Buffer size} & \textbf{Accuracy ($\%$)}\\
    \midrule
    w/o distillation&0&53.59$\pm$0.63\\
    w/ $\mathcal{L}_{IRD}$&0&63.65$\pm$0.55\\
    w/ $\mathcal{L}_{S-PRD}$&0&64.17$\pm$0.41\\
    w/ $\mathcal{L}_{HSD}$&0&\textbf{69.26$\pm$0.32}\\
    \hline
    w/o distillation&200&53.62$\pm$0.81\\
    w/ $\mathcal{L}_{IRD}$&200&70.54$\pm$0.95\\
    w/ $\mathcal{L}_{S-PRD}$&200&69.20$\pm$0.58\\
    w/ $\mathcal{L}_{HSD}$&200&\textbf{72.63$\pm$0.78}\\
    \bottomrule
\end{tabular}
  }
  \caption{Ablation study on the effectiveness of $\mathcal{L}_{HSD}$ in class-IL. We run all tests on the Seq-Cifar-10 dataset with the plasticity loss $\mathcal{L}_{FNC^2}$ and different distillation methods.}
  \label{tab:stability-verification}
\end{table}

\noindent \textbf{Effectiveness of pseudo-replay prototypes.} We conduct ablation experiments using the class-IL setup on the Seq-Cifar-10 and Seq-Cifar-100 datasets with buffer size 0 and 200 to evaluate the role of pseudo-replay prototypes in our method. As shown in \cref{tab:pseudo-replay-verification}, this approach consistently improves performance across all datasets and buffer sizes. Notably, in the memory-free scenario, accuracy with pseudo-replay prototypes is close to that achieved with a buffer size of 200 without them. These outcomes validate our hypothesis that prior prototypes can act as effective representatives of past samples, enhancing cross-task consolidation when used as negative points in current tasks.   
\begin{table}
    \centering
    \resizebox*{!}{0.33\columnwidth}{
        \begin{tabular}{ccccc}
    \toprule
    \multirow{2}{*}{\textbf{Buffer}} & \multirow{2}{*}{\textbf{Pseudo-replay}} & \multicolumn{2}{c}{\textbf{Dataset}}\\
    &&\textbf{Seq-Cifar-10}&\textbf{Seq-Cifar-100} \\
    \midrule
    \multirow{2}{*}{0} & \ding{55}&67.79$\pm$0.43&30.80$\pm$0.49\\
    &\checkmark&\textbf{69.26$\pm$0.32}&\textbf{32.57$\pm$0.55}\\
    \midrule
    \multirow{2}{*}{200} & \ding{55}&70.71$\pm$0.89&33.12$\pm$0.81\\
    &\checkmark&\textbf{72.63$\pm$0.78}&\textbf{34.04$\pm$0.42}\\
    \bottomrule
\end{tabular}
    }
    \caption{Influence of pseudo-replay prototypes.}
    \label{tab:pseudo-replay-verification}
\end{table}

\section{Conclusion}
\label{sec:conclusion}

We explore the roles of hard (``hardness'') and soft (``softness'') relationships in NC-based CL for both \textit{plasticity} and \textit{stability}. To address these, we propose two loss functions: $\mathcal{L}_{FNC^2}$ for plasticity, which uses fixed prototypes to guide representations towards optimal points while emphasizing hard samples and implicitly incorporating \textit{cross-task} consolidation through pseudo-replay of old prototypes, and $\mathcal{L}_{HSD}$ for distilling both hardness and softness over time. Our approach achieves SoTA in memory-free settings across various datasets and remains competitive with limited buffer sizes in memory-based scenarios.

\textbf{Limitations and future work.} Like other NC-inducing methods in CL, our approach is limited by the need to pre-define prototypes, which is impractical when the number of prototypes is unknown. In future work, inspired by recent advancements \cite{lu2024learningmixtureprototypesoutofdistribution}, we will tackle this issue by pre-defining a maximum number of prototypes and learning their distributions, allowing samples to be associated with multiple prototypes and different weights. Additionally, we plan to explore alternative memory-free evaluation methods, as current approaches including Co$^2$L \cite{co2l}, CILA \cite{wen2024provablecontrastivecontinuallearning} rely on buffers, which are less effective with limited samples for old classes. We also aim to identify easily forgotten samples and focus on distilling only the core knowledge.



\section*{Acknowledgements}
This work is partially supported by the ANR-21-ASRO-0003 ROV-Chasseur project.


{\small
\bibliographystyle{ieee_fullname}
\bibliography{egbib}
}

\clearpage
\appendix


\noindent\textbf{APPENDIX}
\section{Hyperparameter selection}
\FloatBarrier 
To select hyperparameters, we employ a grid search strategy, using a randomly drawn $10\%$ of the training data as the validation set. The considered hyperparameters are:
\begin{itemize}
    \item Learning rate ($\eta$)
    \item Batch size (\textit{bsz})
    \item Number of start epochs ($E_1$)
    \item Number of epochs of $t$-th task ($E_{t \ge 2}$)
    \item Temperature for plasticity loss ($\tau$): We use the same $\tau$ for both Focal Neural Collapse Contrastive (FNC$^2$) and Asymmetric SupCon loss \cite{co2l} 
    \item Focusing hyperparameters ($\gamma$) for FNC$^2$ loss
    \item Temperature for instance-wise relation distillation loss ($\mathcal{L}_{IRD}$): As in \cite{co2l}, we use different temperature hyperparameters for the past ($\kappa_{past}$) and current ($\kappa_{current}$) similarity vectors
    \item Temperature for sample-prototype relation distillation loss ($\mathcal{L}_{S-PRD}$): We utilize $\zeta_{past}$ for the past and $\zeta_{current}$ for the current similarity vectors
    \item Number of warm-up epochs in hardness-softness distillation loss ($\mathcal{L}_{HSD}$) ($e_0$)
\end{itemize}

\begin{table}
    \centering
    \resizebox*{!}{0.52\columnwidth}{
        \begin{tabular}{cc}
    \toprule
    \textbf{Hyperparameter} & \textbf{Values} \\
    \midrule
    $\eta$ & \{0.1, 0.5, 1.0\}\\
    \textit{bsz} & \{256, 512\} \\
    $E_1$  & \{500\} \\
    $E_{t \geq 2}$ & \{50, 100\} \\
    $\tau$ & \{0.1, 0.5, 1.0\} \\
    $\gamma$ & \{0, 1, 2, 4, 7, 10\} \\
    $\kappa_{past}$ & \{0.01, 0.05, 0.1\} \\
    $\kappa_{current}$ & \{0.1, 0.2\} \\
    $\zeta_{past}$ & \{0.01, 0.05, 0.1\} \\
    $\zeta_{current}$ & \{0.1, 0.2\} \\
    $e_0$ & \{10, 20, 30\} \\
    \bottomrule
\end{tabular}
    }
    \caption{Search spaces of hyperparameters.}
    \label{tab:search-space-hyperparameters}
\end{table}

The corresponding search space of these hyperparameters are provided in \cref{tab:search-space-hyperparameters}. The selections of these hyperparameters are based on the average test accuracy over five independent trials, and the final chosen values are detailed in \cref{tab:chosen-hyperparameters}. For the sake of conciseness and to maintain focus, we omit those hyperparameters previously discovered in the literature.

\begin{figure*}[t]
  \centering
   \includegraphics[width=1.0\linewidth]{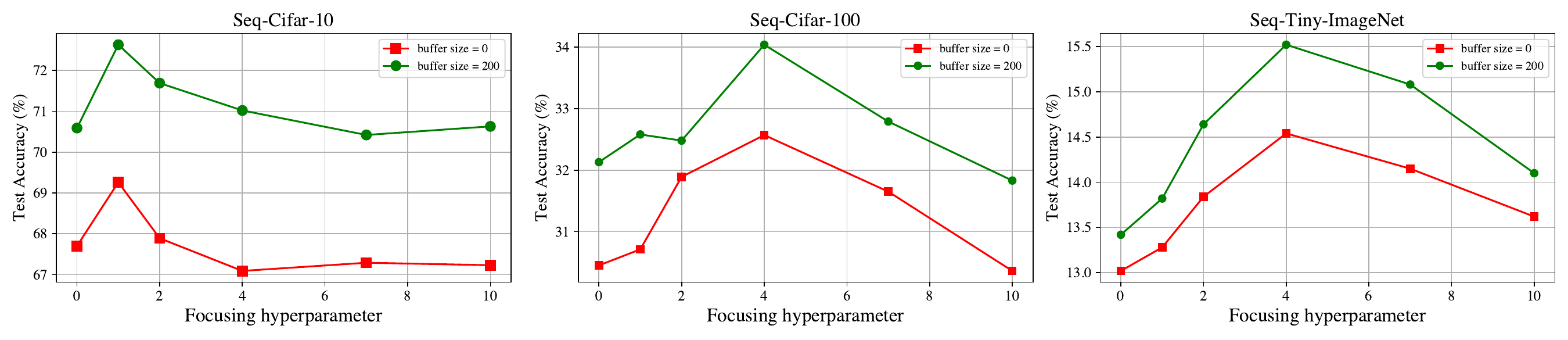}

   \caption{Test accuracy over different values of $\gamma$.}
   \label{fig:gamma-selection}
\end{figure*}

\textbf{Focusing hyperparameter ($\boldsymbol{\gamma}$).} In the FNC$^2$ loss function, $\gamma$ plays a crucial role in determining the level of focus on hard samples (i.e., positive samples that are far from the anchor or their prototypes). To explore this role and select the most suitable $\gamma$ for each dataset, we conduct experiments across different datasets to observe how the the performance of our method changes as $\gamma$ varies. The test accuracy results in \cref{fig:gamma-selection}  show that our method performs best at different values of $\gamma$ for each dataset. Specifically, as reported in \cref{tab:chosen-hyperparameters}, the chosen $\gamma$ for the Seq-Cifar-100 and Seq-Tiny-Imagenet datasets ($\gamma=4$ for both) are larger than that for the Seq-Cifar-10 dataset ($\gamma=1$). This difference arises because Seq-Cifar-100 and Seq-Tiny-ImageNet have a large number of classes per task (both have 20 classes/task), which increases the likelihood of samples being close to the prototypes of other class clusters. In contrast, the Seq-Cifar-10 dataset has only 2 classes each task, making it less complex and not requiring a large $\gamma$.
\begin{table}
    \centering
    \resizebox*{!}{0.45\columnwidth}{
        \begin{tabular}{cccc}
    \toprule
    \textbf{Method} & \textbf{Buffer size} & \textbf{Dataset} & \textbf{Hyperparameter} \\
    \midrule
    \multirow{9}{*}{Our}&\multirow{3}{*}{0, 200, 500}&\multirow{3}{*}{Seq-Cifar-10}&$\eta$: 0.5, $\gamma$: 1, \textit{bsz}: 512, $E_1$: 500, $E_{t \ge 2}$: 100, \\
    &&&$\tau$: 0.5, $e_0$: 30, $\kappa_{past}$: 0.01, $\kappa_{current}$: 0.2,\\
    &&&$\zeta_{past}$: 0.01, $\zeta_{current}$: 0.2 \\
    
    \cline{2-4} 
    
    & \multirow{3}{*}{0, 200, 500} & \multirow{3}{*}{Seq-Cifar-100}&$\eta$: 0.5, $\gamma$: 4, \textit{bsz}: 512, $E_1$: 500, $E_{t \ge 2}$: 100, \\
    &&&$\tau$: 0.5, $e_0$: 30, $\kappa_{past}$: 0.01, $\kappa_{current}$: 0.2, \\
    &&&$\zeta_{past}$: 0.1, $\zeta_{current}$: 0.2 \\

    \cline{2-4} 
    & \multirow{3}{*}{0, 200, 500} & \multirow{3}{*}{Seq-Tiny-ImageNet}&$\eta$: 0.1, $\gamma$: 4, \textit{bsz}: 512, $E_1$: 500, $E_{t \ge 2}$: 50, \\
    &&&$\tau$: 0.5, $e_0$: 20, $\kappa_{past}$: 0.1, $\kappa_{current}$: 0.1, \\
    &&&$\zeta_{past}$: 0.1, $\zeta_{current}$: 0.2 \\
    \midrule
    \multirow{4}{*}{Co$^2$L}&\multirow{2}{*}{0, 200, 500} &\multirow{2}{*}{Seq-Cifar-100}&$\eta: 0.5$, \textit{bsz}: 512, $E_1$: 500, $E_{t \ge 2}$: 100, \\
    &&& $\tau$: 0.5, $\kappa_{past}$: 0.01, $\kappa_{current}$: 0.2 \\
    \cline{2-4}
    & \multirow{2}{*}{0} &\multirow{2}{*}{Seq-Tiny-ImageNet}&$\eta: 0.1$, \textit{bsz}: 512, $E_1$: 500, $E_{t \ge 2}$: 50, \\
    &&&$\tau$: 0.5, $\kappa_{past}$: 0.1, $\kappa_{current}$: 0.1 \\
    \bottomrule
\end{tabular}

    }
    \caption{Selected hyperparameters in our experiments.}
    \label{tab:chosen-hyperparameters}
\end{table}



\section{Additional Experiments}
\label{sec:additional-experiments}
\subsection{Average accuracy results with buffer size 500}
In addition to the results with small buffer sizes (0 and 200), we run experiments with a buffer size of 500 across different datasets to further assess the effectiveness of our method with a larger buffer. As shown in \cref{tab:acc-result-500}, although our method does not surpass state-of-the-art methods, it achieves results close to them on Seq-Cifar-10 and Seq-Tiny-ImageNet, underperforming only on Seq-Cifar-100 compared to GCR \cite{tiwari2022gcrgradientcoresetbased}. This further demonstrates that our method, aside from excelling in memory-free and small buffer settings, remains effective with larger buffers. 

\begin{table*}
    \centering
    \resizebox*{!}{0.48\columnwidth}{
        \begin{tabular}{clccccccc}
\hline

\multirow{2}{*}{\textbf{Buffer}}&\textbf{Dataset}&\multicolumn{2}{c}{\textbf{Seq-Cifar-10}}&\multicolumn{2}{c}{\textbf{Seq-Cifar-100}}&\multicolumn{2}{c}{\textbf{Seq-Tiny-ImageNet}}\\
            &\textbf{Scenario}&\textbf{Class-IL}&\textbf{Task-IL}&\textbf{Class-IL}&\textbf{Task-IL}&\textbf{Class-IL}&\textbf{Task-IL}\\
\hline
    \multirow{9}{*}{500}
    &ER \cite{riemer2019learninglearnforgettingmaximizing}&57.74$\pm$0.27&93.61$\pm$0.27&27.66$\pm$0.61&66.23$\pm$1.52&9.99$\pm$0.29&48.64$\pm$0.46\\
    &iCaRL \cite{rebuffi2017icarl}&47.55$\pm$3.95&88.22$\pm$2.62&33.25$\pm$1.25&58.16$\pm$1.76&9.38$\pm$1.53&31.55$\pm$3.27\\
    &GEM \cite{lopezpaz2017gem}&26.20$\pm$1.26&92.16$\pm$0.64&25.54$\pm$0.65&66.31$\pm$0.86&-&-\\
    &GSS \cite{aljundi2019gradientbasedsampleselection}&49.73$\pm$4.78&91.02$\pm$1.57&21.92$\pm$0.34&60.28$\pm$1.18&-&-\\
    &DER \cite{buzzega2020darkexperiencegeneralcontinual}&70.51$\pm$1.67&93.40$\pm$0.39&41.36$\pm$1.76&71.73$\pm$0.74&17.75$\pm$1.14&51.78$\pm$0.88\\
    &Co$^2$L \cite{co2l}&74.26$\pm$0.77&95.90$\pm$0.26&37.02$\pm$0.76&62.44$\pm$0.36&20.12$\pm$0.42&53.04$\pm$0.69\\
    &GCR \cite{tiwari2022gcrgradientcoresetbased}&74.69$\pm$0.85&94.44$\pm$0.32&\textbf{45.91$\pm$1.30}&\textbf{71.64$\pm$2.10}&19.66$\pm$0.68&52.99$\pm$0.89\\
    &CILA \cite{wen2024provablecontrastivecontinuallearning}&\textbf{76.03$\pm$0.79}&\textbf{96.40$\pm$0.21}&-&-&\textbf{20.64$\pm$0.59}&\textbf{54.13$\pm$0.72}\\
    &\text{Ours}&75.51$\pm$0.52&96.14$\pm$0.25&{40.25$\pm$0.58}&65.85$\pm$0.44&{20.31$\pm$0.34}&{53.46$\pm$0.59}\\
   
\hline

\end{tabular}
    }
    \caption{Additional results with buffer size 500 (best results in each column are bold).}
    \label{tab:acc-result-500}
\end{table*}

\subsection{Average forgetting results}
We utilize the Average Forgetting metric as defined in \cite{chaudhry_2018} to quantify how much information the model has forgotten about previous tasks, which as
\begin{equation}
    \label{eq:forgetting-metric}
    F = \frac{1}{T-1}\sum_{i=1}^{T-1}{max_{t \in \{1,\ldots,T-1\}}{(A_{t,i}-A_{T,i})}}
\end{equation}
\cref{tab:forgetting-result} report the average forgetting results of our method compared to all other baselines. The results show that our method can effectively mitigate forgetting, especially even without using additional buffers.
\begin{table*}
    \centering
    \resizebox*{!}{0.95\columnwidth}{
        \begin{tabular}{clccccccc}
\hline
\multirow{2}{*}{\textbf{Buffer}}&\textbf{Dataset}&\multicolumn{2}{c}{\textbf{Seq-Cifar-10}}&\multicolumn{2}{c}{\textbf{Seq-Cifar-100}}&\multicolumn{2}{c}{\textbf{Seq-Tiny-ImageNet}}\\
&\textbf{Scenario}&\textbf{Class-IL}&\textbf{Task-IL}&\textbf{Class-IL}&\textbf{Task-IL}&\textbf{Class-IL}&\textbf{Task-IL}\\

\hline
    {0}
    &{Co$^2$L}\cite{co2l}&35.81$\pm$1.08&14.33$\pm$0.87&66.51$\pm$0.28&39.63$\pm$0.62&62.80$\pm$0.77&39.54$\pm$1.08\\

    &{\text{Ours}}&\textbf{23.85$\pm$0.30}&\textbf{4.72$\pm$0.28}&\textbf{52.03$\pm$0.63}&\textbf{36.20$\pm$0.48}&\textbf{53.97$\pm$0.63}&\textbf{37.57$\pm$0.88}\\
\hline
    {200}
    &ER \cite{riemer2019learninglearnforgettingmaximizing} &59.30$\pm$2.48&6.07$\pm$1.09&75.06$\pm$0.63&	27.38$\pm$1.46	&76.53$\pm$0.51&	40.47$\pm$1.54
     \\
    &GEM \cite{lopezpaz2017gem}&80.36$\pm$5.25&9.57$\pm$2.05&77.40$\pm$1.09&29.59$\pm$1.66
    &-&- \\ 
    &GSS \cite{aljundi2019gradientbasedsampleselection}&72.48$\pm$4.45&	8.49$\pm$2.05&	77.62$\pm$0.76&	32.81$\pm$1.75&	- &	-
     \\
    &iCARL \cite{rebuffi2017icarl}&\textbf{23.52$\pm$1.27}&	25.34$\pm$1.64&\textbf{47.20$\pm$1.23}&36.20$\pm$1.85&	\textbf{31.06$\pm$1.91}&	42.47$\pm$2.47
     \\
    &DER\cite{buzzega2020darkexperiencegeneralcontinual}&35.79$\pm$2.59& 6.08$\pm$0.70&62.72$\pm$2.69&25.98$\pm$1.55 &64.83$\pm$1.48&40.43$\pm$1.05\\
    &Co$^2$L \cite{co2l}&36.35$\pm$1.16& 6.71$\pm$0.35
    &67.82$\pm$0.41&38.22$\pm$0.34
     &73.25$\pm$0.21&47.11$\pm$1.04
    \\
    &GCR\cite{tiwari2022gcrgradientcoresetbased}&32.75$\pm$2.67& 7.38$\pm$1.02&57.65$\pm$2.48&\textbf{24.12$\pm$1.17}&65.29$\pm$1.73&40.36$\pm$1.08\\

    &{CILA}\cite{wen2024provablecontrastivecontinuallearning}&-&-&-&-&-&-\\
    
    &{\text{Ours}}&25.24$\pm$0.69&\textbf{4.28$\pm$0.32}&52.40$\pm$0.83&33.66$\pm$0.24&52.07$\pm$0.46&\textbf{33.76$\pm$0.58}\\
\hline
    {500}
    &ER \cite{riemer2019learninglearnforgettingmaximizing} &43.22$\pm$2.10&3.50$\pm$0.53&67.96$\pm$0.78&	17.37$\pm$1.06	&75.21$\pm$0.54&	30.73$\pm$0.62
     \\
    &GEM \cite{lopezpaz2017gem}&78.93$\pm$6.53&5.60$\pm$0.96&71.34$\pm$0.78&	20.44$\pm$1.13&-&-
     \\
    &GSS \cite{aljundi2019gradientbasedsampleselection}&59.18$\pm$4.00&	6.37$\pm$1.55&	74.12$\pm$0.42&	26.57$\pm$1.34&	- & -
     \\
    &iCARL \cite{rebuffi2017icarl}&28.20$\pm$2.41&	22.61$\pm$3.97&	40.99$\pm$1.02&	27.90$\pm$1.37	&\textbf{37.30$\pm$1.42}	&39.44$\pm$0.84
     \\
    &DER\cite{buzzega2020darkexperiencegeneralcontinual}&24.02$\pm$1.63&3.72$\pm$0.55 &49.07$\pm$2.54& 25.98$\pm$1.55&59.95$\pm$2.31&28.21$\pm$0.97\\
    
    &Co$^2$L \cite{co2l}&25.33$\pm$0.99&3.41$\pm$0.80
     &51.23$\pm$0.65&26.30$\pm$0.57
     &65.15$\pm$0.26&39.22$\pm$0.69
    \\
    
    &GCR\cite{tiwari2022gcrgradientcoresetbased}&\textbf{19.27$\pm$1.48}&\textbf{3.14$\pm$0.36} &\textbf{39.20$\pm$2.84}& \textbf{15.07$\pm$1.88}&56.40$\pm$1.08&27.88$\pm$1.19\\

    &{CILA}\cite{wen2024provablecontrastivecontinuallearning}&-&-&-&-&-&-\\
    
    &{\text{Ours}}&22.59$\pm$1.02&3.21$\pm$0.25&41.66$\pm$0.78&24.84$\pm$0.91&46.08$\pm$0.56&\textbf{26.45$\pm$0.79}\\
\hline
\end{tabular}
    }
    \caption{Average forgetting (lower is better) across five independent trials: Comparison of our method with all baselines in continual learning.}
    \label{tab:forgetting-result}
\end{table*}



\end{document}